\documentclass{article}

\usepackage{arxiv,lineno,microtype}

\usepackage[utf8]{inputenc} 
\usepackage[T1]{fontenc}    
\usepackage{hyperref}       
\usepackage{url}            

\usepackage{booktabs} 
\usepackage{graphicx}
\usepackage{dsfont}
\usepackage{amssymb}
\usepackage{amsmath}
\usepackage{color}
\usepackage{subcaption}
\usepackage{enumitem}
\usepackage{authblk}

\title{Are Nearby Neighbors Relatives?: Testing Deep Music Embeddings\thanks{this work was accepted for publication in the "Frontiers in Applied Mathematics and Statistics (Deep Learning: Status, Applications and Algorithms)"}}

\author[1, $\dagger$]{\small Jaehun Kim}
\author[1]{\small Juli\'{a}n Urbano}
\author[1]{\small Cynthia C. S. Liem}
\author[1]{\small Alan Hanjalic}

\affil[1]{\footnotesize Multimedia Computing Group, Delft University of Technology, Delft, the Netherlands}
\affil[$\dagger$]{\footnotesize Correspondance, \texttt{j.h.kim@tudeleft.nl}}

\begin{document}
\maketitle

\begin{abstract}
Deep neural networks have frequently been used to directly learn representations useful for a given task from raw input data. In terms of overall performance metrics, machine learning solutions employing deep representations frequently have been reported to greatly outperform those using hand-crafted feature representations. At the same time, they may pick up on aspects that are predominant in the data, yet not actually meaningful or interpretable. 
In this paper, we therefore propose a systematic way to test the trustworthiness of deep music representations, considering musical semantics. The underlying assumption is that in case a deep representation is to be trusted, distance consistency between known related points should be maintained both in the input audio space and corresponding latent deep space. We generate known related points through semantically meaningful transformations, both considering imperceptible and graver transformations. Then, we examine within- and between-space distance consistencies, both considering audio space and latent embedded space, the latter either being a result of a conventional feature extractor or a deep encoder.
We illustrate how our method, as a complement to task-specific performance, provides interpretable insight into what a network may have captured from training data signals.
\end{abstract}

\keywords{music information retrieval \and neural network \and representation learning \and evaluation \and MFCC}

\section{Introduction}\label{sec:introduction}

Music audio is a complex signal. Frequencies in the signal usually belong to multiple pitches, which are organized harmonically and rhythmically, and often originate from multiple acoustic sources in the presence of noise. When solving tasks in the Music Information Retrieval (MIR) field, within this noisy signal, the optimal subset of information needs to be found that leads to quantifiable and musical descriptors. Commonly, this process is handled by pipelines exploiting a wide range of signal processing and machine learning algorithms.
Beyond the use of \textit{hand-crafted music representations}, which are informed by human domain knowledge, as an alternative, \textit{deep music representations} have emerged, that are trained by employing deep neural networks (DNNs) and massive amounts of training data observations. Such deep representations are usually reported to outperform hand-crafted representations (e.g. \cite{NIPS2013_5004, humphrey2012moving, choi2017convolutional, chandna2017monoaural}).

\begin{figure}
  \centering
  \includegraphics[width=0.7\textwidth]{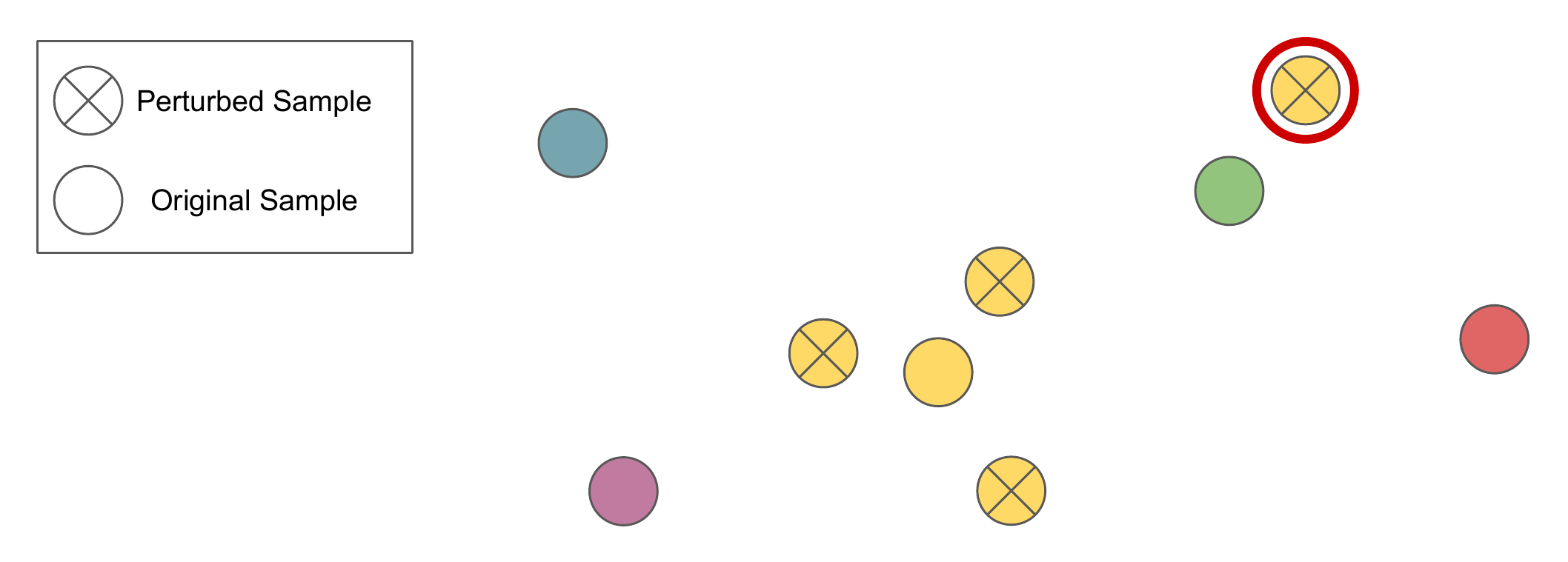}
  \caption{Simplified example illustrating distance assumption within a space. Circles without a cross indicate music clips. Yellow circles with crosses refer to hardly perceptible transformations of the yellow original clip. The top-right transformation, marked with a red outer circle, actually is closer to another original clip (green) than to its own original (yellow), which violates the assumption it should be closest to its original, and hence may be seen as an error-inducing transformation under a nearest-neighbor scheme.}
  \label{fig:example}
\end{figure}

At the same time, the \textit{performance} of MIR systems may be vulnerable to subtle input manipulation. The addition of small noise may lead to unexpected random behavior, regardless of whether traditional or deep models are used \cite{DBLP:journals/tmm/Sturm14,DBLP:conf/ismir/Rodriguez-Algarra16,DBLP:journals/cie/Sturm16,DBLP:journals/tmm/KereliukSL15}. In a similar line of thought, in the broader deep learning (DL) community, increasing attention is given to adversarial examples that are barely differentiable from original samples, but greatly impact a network's performance~\cite{DBLP:journals/corr/GoodfellowSS14, DBLP:journals/tmm/KereliukSL15}. 

So far, the sensitivity of representations with respect to subtle input changes has mostly been tested in relation to dedicated machine learning tasks (e.g.\ object recognition, music genre classification), and examined by investigating whether these input changes cause performance drops. When purely considering the questions \emph{whether relevant input signal information can automatically be encoded into a representation, and to what extent the representation can be deemed `reliable'}, in principle, the learned representation should be general and useful to different types of tasks. Therefore, in this work, we will not focus on performance obtained by using a learned representation for certain machine learning tasks, but rather on a systematic way to verify assumptions on distance relationships between several representation spaces: the audio space and the learned space.

Inspired by~\cite{DBLP:journals/tmm/Sturm14}, we will also investigate the effect of musical and acoustic transformations of audio input signals, in combination with an arbitrary encoder of the input signal, which either may be a conventional feature extractor or deep learning-based encoder. In doing this, we have the following major assumptions:

\begin{enumerate}[label=(\roman*)\hspace{0.2cm}, itemindent=2em]
    \item if a small, humanly imperceptible transformation is introduced, the distance between the original and transformed signal should be very small, both in the audio and encoded space. This is illustrated in Figure \ref{fig:example}
    \item however, if a more grave transformation is introduced, the distance between the original and transformed signal should be larger, both in the audio and encoded space.
    \item the degree of how these assumptions hold will differ for the tasks and the datasets on which the encoder is trained.
\end{enumerate}

To examine the above assumptions, we seek to answer the following research questions:

\begin{enumerate}[label=RQ \arabic*.\hspace{0.2cm}, itemindent=4em]
    \item Do assumption (i) and (ii) hold for conventional and deep learning-based encoders?
    \item Does assumption (iii) hold for music-related tasks and corresponding datasets, especially when deep learning is applied?
\end{enumerate}

By answering the above questions, ultimately we seek to test if considered music-related encoders hold a desirable consistency, such that the distances between audio space and the latent space are monotonically related.

With this work, we intend to offer directions towards a complementary evaluation method for deep machine learning pipelines, that focuses on space diagnosis rather than the troubleshooting of pipeline output. Our intention is that this will provide the researcher with additional insight into the reliability and potential semantic sensitivities of deep learned spaces.

In the remainder of this paper, we first describe our approaches including the details on the learning setup (Section \ref{sec:learning}) and the methodology to assess distance consistency (Section \ref{sec:measuring}), followed by the experimental setup (Section \ref{sec:experiment}). Further, we report the result from our experiments (Section \ref{sec:result}). Afterwards we discuss the results and conclude this work (Section \ref{sec:disc_concl}).

\section{Learning}\label{sec:learning}

To diagnose a deep music representation space, such a space should first exist. For this, one needs to find a learnable deep encoder $f:\mathbb{R}^{t\times{b}}\to\mathbb{R}^{d}$ that transforms the input audio representation $x\in\mathbb{R}^{t\times{b}}$ to a latent vector $z\in\mathbb{R}^{d}$, while taking into account the desired output for a given learning task. The learning of $f$ can be done by adjusting the parametrization $\Theta^{f}$ to optimize the objective function, which should be defined in accordance to a given task.

\subsection{Tasks}\label{subsec:task}

In our work, we consider representations learned for four different tasks: \emph{autoencoder} (AE), music \emph{auto-tagging} (AT), \emph{predominant instrument recognition} (IR), and finally singing \emph{voice separation} (VS). By doing this, we take a broad range of problems into account that are particularly common in the MIR field. AE is a representative task for unsupervised learning using DNNs, and AT is a popular supervised learning task in the MIR field~\cite{choi2017convolutional,DBLP:conf/ismir/ChoiFSC17,DBLP:journals/corr/abs-1712-00866,DBLP:journals/corr/abs-1710-10451,DBLP:conf/icassp/DielemanS14, app8010150}. AT is a multi-label classification problem, in which individual labels are not always mutually exclusive and often highly inter-correlated. As such, it can be seen as a more challenging problem than IR, which is a single-label classification problem. Furthermore, IR labels involve instruments, which can be seen as more objective and taxonomically stable labels than e.g.\ genres or moods. Finally, VS is a task that can be formulated as a regression problem, that learns a mask to segregate a certain region of interest out of a given signal mixture.

\subsubsection{Autoencoder}\label{subsec:ae}
The objective of an autoencoder is to find a set of encoder $f$ and decoder $g$ functions such that the input audio $x$ is encoded into a fixed-length vector and reconstructed as follows:

\begin{equation}\label{eq:ae_model}
    \hat{x} = g(f(x))
\end{equation}

Here, the $\hat{x}=g(f(x))$ is the output of a cascading pipeline of a decoder $g:\mathbb{R}^{d}\to\mathbb{R}^{t\times{b}}$ parameterized by $\Theta^{g}$, followed by an encoder $f$. To obtain a desired model, a reconstruction error is typically considered as its loss function:

\begin{equation}\label{eq:ae_loss}
J^{AE} = \sum_{i=1}^{\vert\mathcal{X}^{tr}\vert}\lVert x^{i} - \hat{x}^{i}\rVert_{2}
\end{equation}

where $\mathcal{X}_{tr}$ is the given set of training samples for the autoencoder task.

\subsubsection{Music Auto-Tagging}\label{subsec:at}

Unlike the autoencoder, a DNN model architecture for either multi-label or multi-class classification has architectural block $h$ to infer the posterior distribution of classes from the encoding by $f$:

\begin{equation}\label{eq:at_model}
    \hat{y} = \sigma(h(f(x)))
\end{equation}

Since we consider a single fully-connected layer as $h$ in this study, $h:\mathbb{R}^{d}\to\mathbb{R}^{K}$ is the prediction layer parameterized by $\Theta_{h}$, which transforms the deep representation $z^{i}$ into the logit per class, which is finally mapped into $p(k|x^{i})$ by the sigmoid function $\sigma$. 

The typical approach to music auto-tagging using DNNs is to consider the problem as a multi-label classification problem, for which the objective is to minimize the binary cross-entropy of each music tag $k\in{\{1, 2, ... , K\}}$, which is expressed as follows:

\begin{equation}\label{eq:at_loss}
J^{AT} = -\sum_{i=1}^{\vert\mathcal{X}^{tr}\vert}\sum_{k=1}^{K} y^{i}_{k}\log{(\hat{y}^{i}_{k})} + (1-y^{i}_{k})\log{(1-\hat{y}^{i}_{k})}
\end{equation}

\noindent where $y^{i}_{k}$ is the binary label that indicates whether the tag $k$ is related to the input audio signal $x^{i}$. Similarly, $\hat{y}^{i}_{k}$ indicates the inferred probability of $x^{i}$ and tag $k$. The optimal functions $f$ and $h$ are found by adjusting $\Theta^{f}$ and $\Theta^{h}$ such that (\ref{eq:at_loss}) is minimized.

\subsubsection{Predominant Musical Instrument Recognition}\label{subsec:ir}

The learning of the IR task can be formulated as a single-label, multi-class classification, which allows one to use a model architecture similar to the aforementioned one, except the terminal non-linearity:

\begin{equation}\label{eq:ir_model}
    \hat{y} = softmax(h(f(x)))
\end{equation}

Here, the softmax function $softmax(o_t) = \frac{e^{o_{t}}}{\sum_{c=1}^{T} e^{o_{c}}}$ , where $o\in\mathbb{R}^{T}$ is the output of $h$, substitutes the sigmoid function in (\ref{eq:at_model}) to output the categorical distribution over the class.

To maximize the classification accuracy, one of the popular loss function especially in the context of neural network learning is categorical cross-entropy, given as follows:

\begin{equation}\label{eq:ir_loss}
J^{IR} = -\sum_{i=1}^{\vert\mathcal{X}^{tr}\vert}\sum_{t=1}^{T} y^{i}_{t}\log{(\hat{y}^{i}_{t})}
\end{equation}

where $t\in\{1,2, ... ,T\}$ is a instrument class and thus, $y_{t}^{i}$ is the binary label of instance $x^{i}$ to the class $t$ and $\hat{y}^{i}_{t}$ indicates the inferred probability of $x^{i}$ and instrument $t$, respectively.

\subsubsection{Singing Voice Separation}\label{subsec:vs}

There are multiple ways to set up an objective function for the source separation task. It can be achieved by simply applying (\ref{eq:ae_loss}) between the output of the network $\hat{x}=g(f(x))$ and the desired isolated signal $s\in\mathcal{R}^{t\times{b}}$ such that the model can infer direct isolated sound. In this case, the objective function is similar to (\ref{eq:ae_loss}), except that the target is substituted from the input signal $x$ to the isolated signal $s$. On the other hand, as introduced in~\cite{DBLP:conf/ismir/JanssonHMBKW17}, one can learn a model predicting the mask that segments the target component from the mixture as follows:

\begin{equation}\label{eq:ss_model}
    \hat{s} = \sigma(g(f(x))) \odot x
\end{equation}

where $\hat{s}$ is the estimated isolated signal and $x\in\mathcal{R}^{t\times{b}}$ is the representation of the original input mixture, and $\odot$ refers to the element-wise multiplication. $\sigma(g(f(x)))\in\mathcal{R}^{t\times{b}}$ is the mask inferred by $g$ and $f$ of which the elements are bounded in the range $[0, 1]$ by the sigmoid function $\sigma$, such that they can be used for the separation of the target source. As introduced in \cite{DBLP:conf/ismir/JanssonHMBKW17}, we applied the skip connections.

For the optimization of the encoder parameters $\Theta_{f}$ and the decoder parameters $\Theta_{g}$, \cite{DBLP:conf/ismir/JanssonHMBKW17} suggests to use the L1 loss as follows:

\begin{equation}\label{eq:vs_loss}
J^{VS} = \sum_{i=1}^{\vert\mathcal{X}^{tr}\vert}\lVert s^{i} - \hat{s}^{i}\rVert_{1}
\end{equation}

where $s^{i}$ is the low-level representation of the isolated signal, which serves as the regression target. Note, that both input $x^i$ and estimated target source $\hat{s}$ are magnitude spectra, so we use the original phase of input $x^i$ to reconstruct a time-domain signal.

\subsection{Network Architectures}\label{subsec:net_arch}

The architecture of a DNN determines the overall structure of the network, which defines the details of the desired patterns to be captured by the learning process~\cite{DBLP:books/daglib/0040158}. In other words, it reflects the way in which a network should \textit{interpret} a given input data representation. In this work, we use a \textit{VGG-like} architecture, one of the most popular and general architectures frequently employed in the MIR field.


\begin{table}[!t]
\caption{Employed network architectures. A decoder $g$ is constructed reversing the layers: convolution (Conv) and fully-connected (FC) layers are transposed, and pooling layers repeat the maximum input values in the pooling window.}
\label{tab:architectures}
    \centering
    \begin{tabular}{ll}
    Layers & Output shape \\ \hline\hline
    Input & $1\!\times\!128\!\times\!512$\\ \hline
    
    Conv $3\!\times\!3$, BN, ReLU  & $16\!\times\!128\!\times\!512$     \\ 
    MaxPooling $2\!\times\!2$ & $16\!\times\!64\!\times\!256$\\  \hline
    
    Conv $3\!\times\!3$, BN, ReLU  & $32\!\times\!64\!\times\!256$     \\ 
    MaxPooling $2\!\times\!2$ & $32\!\times\!32\!\times\!128$\\  \hline
    
    Conv $3\!\times\!3$, BN, ReLU  & $64\!\times\!16\!\times\!64$     \\ 
    MaxPooling $2\!\times\!2$ & $64\!\times\!8\!\times\!32$\\  \hline
    
    Conv $3\!\times\!3$, BN, ReLU  & $128\!\times\!8\!\times\!32$     \\ 
    MaxPooling $2\!\times\!2$ & $128\!\times\!4\!\times\!16$\\  \hline
    
    Conv $3\!\times\!3$, BN, ReLU & $256\!\times\!4\!\times\!16$       \\ 
    MaxPooling $2\!\times\!2$& $256\!\times\!2\!\times\!8$\\ \hline

    Conv $3\!\times\!3$, BN, ReLU & $256\!\times\!2\!\times\!8$       \\ 
    MaxPooling $2\!\times\!2$& $256\!\times\!1\!\times\!4$\\ \hline
    
    GlobalAveragePooling & $256$\\ \hline
    \end{tabular}
\end{table}

The \emph{VGG-like} architecture is a Convolutional Neural Network (CNN) architecture introduced by ~\cite{DBLP:conf/nips/KrizhevskySH12,DBLP:journals/corr/SimonyanZ14a}, which employs tiny rectangular filters. Successes of VGG-like architectures have not only been reported for computer vision tasks, but also in various MIR fields~\cite{choi2017convolutional,DBLP:journals/tmm/KereliukSL15}. The detailed architecture design used in our work can be found in the Table \ref{tab:architectures}.

\subsection{Architecture and Learning Details}\label{subsec:opt}

For both architectures, we used Rectified Linear Units (ReLU)~\cite{Nair2010RectifiedMachines} for the nonlinearity, and Batch Normalization (BN) in every convolutional and fully-connected layer for fast training and regularization~\cite{DBLP:conf/icml/IoffeS15}.
We use Adam~\cite{JHKM:conf/iclr/KingmaB15} as optimization algorithm during training, where the learning rate is set for $0.001$ across all models. We trained models with respect to their objective function, which requires different optimization strategies. Nonetheless, we regularized the other factors except the number of epochs per task, which inherently depends on the dataset and the task. The termination point of the training is set manually, where either the validation loss reaches to the plateau or starts to increase. More specifically, we stopped the training for each task at the epoch of $\{500, 200, 500, 5000\}$ for the AE, AT, IR, VS task, respectively.

\begin{figure}
    \centering
    \includegraphics[width=1\linewidth]{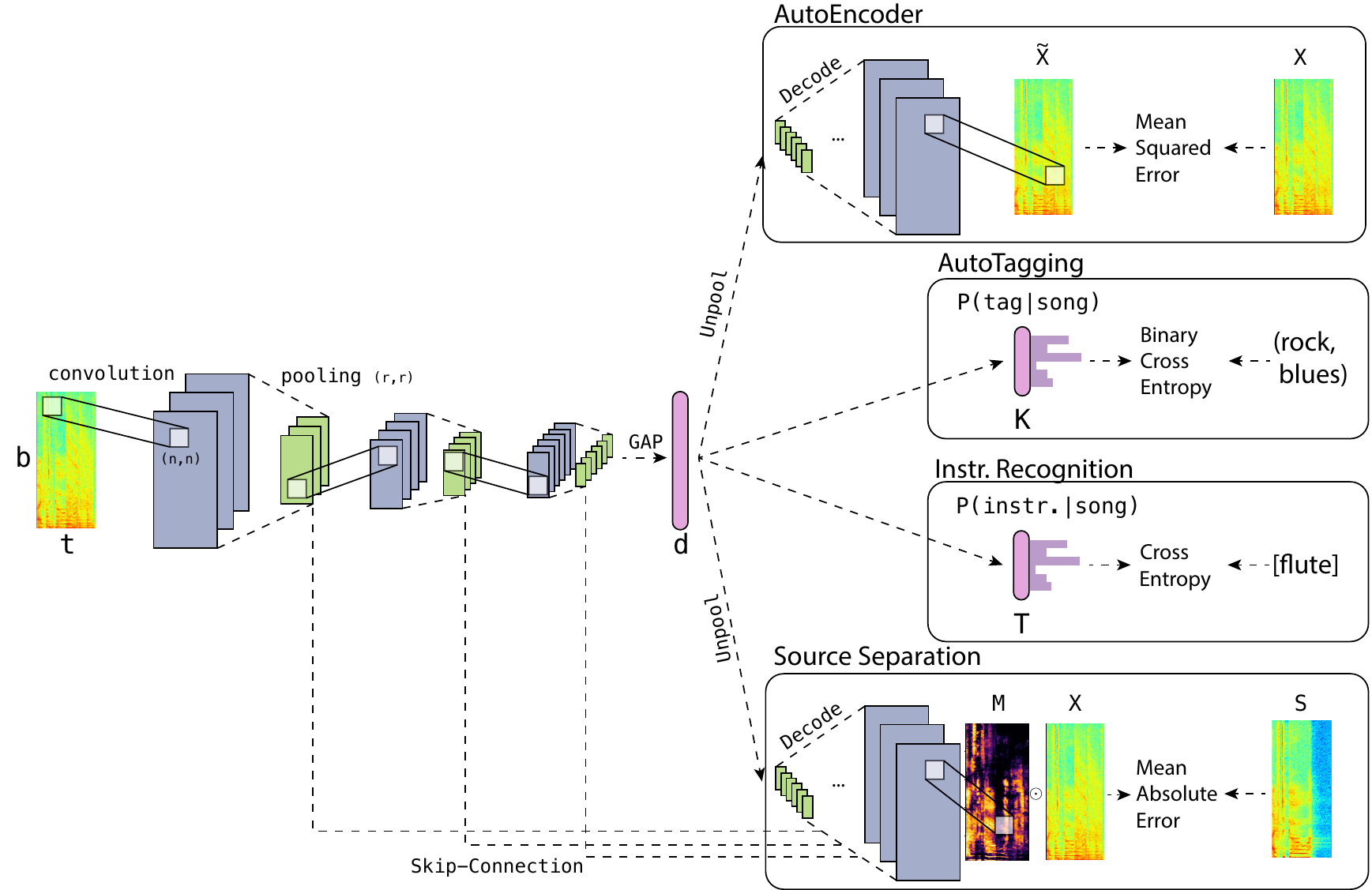}
    \caption{Network architecture used in this work. The left half of the model is the encoder pipeline $f$, whose architecture is kept the same across all the tasks of our experiments. The pink vertical bar represents the latent vector $z$, in which all the measures we propose are tested. The right half of the diagram refers to the four different prediction pipelines with respect to the tasks. The top block describes the decoder and the error function of the task (where, for simplicity, detailed illustrations of decoder $g$ of $f$ are omitted). The second and third block represent the AT and IR task, respectively. Here, the smaller pink bar represents the terminal layer for the prediction of the posterior distribution for $K$ music tags or $T$ musical instruments. Finally, the lowest block is describing the mask predictor $g$, prediction process and the way the error function is calculated. Also, this architecture includes the skip-connections from each convolution block of the encoder, which is the key characteristic of the U-Net~\cite{DBLP:conf/miccai/RonnebergerFB15}.}
    \label{fig:arch}
\end{figure}

\section{Measuring Distance Consistency}\label{sec:measuring}

In this work, among the set of potential representation spaces, we consider two specific subsets of representation spaces of interest: the audio input space and the latent embedding space. Let $\mathcal{A}$ be the space where the low-level audio representation of music excerpts belong to. $\mathcal{X}\subset{\mathcal{A}}$ is the set of music excerpts in the dataset and $x\in{\mathcal{X}}$ is each instance. Likewise, $\mathcal{L}$ is the latent space where the set of latent points ${z}\in\mathcal{Z}\subset{\mathcal{L}}$ belongs to. Therefore, an encoder  $f:\mathcal{A}\to\mathcal{L}$ is trained on task-specific training data $\mathcal{X}$ and maps points from $\mathcal{X}$ to $\mathcal{Z}$ while it actually maps $\mathcal{A}$ to $\mathcal{L}$. Specifically, a fixed number of latent spaces per task $\{\mathcal{L}_{AE}, \mathcal{L}_{AT}, \mathcal{L}_{IR}, \mathcal{L}_{VS}\}$ are considered. For all relevant encoders, we will assess their reliability by examining the distance consistency with respect to a set of transformations\footnote{Note that, the term `transformation' differs from the `maps', which correspond to encoders $f$ in our study. While It is rather close to the concept of `input perturbation' from literature, we intentionally avoid using the term, since we also study more grave ranges of deformations which are not usually studied.} $\mathcal{T}=\{\tau_{l}:\mathcal{A}\to\mathcal{A}, l\in[1, 2, ..., L]\}$ and a set of testing points $\mathcal{X}^{ts}\subset \mathcal{A}$. 

In Section~\ref{subsec:distance_consistency}, we describe how distance consistency will be measured. Section~\ref{subsec:distance} will discuss the distance measures that will be used, while Section~\ref{subsec:transformations} discusses what transformations will be adopted in our experiments.

\subsection{Distance Consistency}\label{subsec:distance_consistency}

For distance consistency, we will compute \textit{within-space consistency} and \textit{between-space consistency}.

\subsubsection{Within-Space Consistency}\label{subsubsec:within_space}

For all audio samples $x\in\mathcal{X}^{ts}$ and transformations $\tau\in\mathcal{T}$, we obtain the transformed points $x_{\tau}=\tau(x)$ and $z_{\tau} = f(x_{\tau})$ first, and then we calculate the error function $\delta$ of each transformed sample as follows:

\begin{equation}\label{eq:fault}
\delta(p, \mathcal{P}, \tau, d)=
\begin{cases}
    0, & \text{if } d(p_{\tau}, p) < d(p_{\tau}, p^{\prime}), \forall{p^{\prime}}\in{\mathcal{P}}\setminus{p}\\
    1 & \text{otherwise}
\end{cases}
\end{equation}

Where $p\in\mathcal{P}$ can be either audio samples $x\in\mathcal{X}$ or latent points $z\in\mathcal{Z}$, according to the target space to be measured. Finally, $d$ is a distance function between two objects.

As $\delta$ indicates how the space is unreliable at the exemplar-level, the within-space consistency can be defined as the complement of $\delta$:

\begin{equation}
C^{W} = 1-\mathbb{E}_{p\in\mathcal{P}}[\delta(p, \mathcal{P}, \tau, d)]
\end{equation}

\subsubsection{Between-Space Consistency}\label{subsubsec:between_space}

To measure consistency between the associated spaces, one can measure how they are correlated. The distances between a transformed point $p_{\tau}$ and its original sample $p$ will be used as characteristic information to make comparisons between spaces. As mentioned above, we consider two specific spaces: the audio input space $\mathcal{A}$ and the embedding space $\mathcal{L}$. Consequently, we can calculate the correlation of distances for the points belonging to each subset of spaces as follows:

\begin{equation}
C^{B}_{\rho} = \rho(d_{\mathcal{A}}^{\tau}, d_{\mathcal{L}}^{\tau})
\end{equation}

\noindent where $\rho$ is Spearman's rank correlation, and $d_{\mathcal{A}}^{\tau}$ and $d_{\mathcal{L}}^{\tau}$ refers to the distance array $d(x_{\tau}, x^{\prime})$ and $d(z_{\tau}, z^{\prime}), \forall{x^{\prime}}\in{\mathcal{X}^{ts}}\setminus{x}$, respectively.

On the other hand, one can also simply measure the agreement between distances, which is given by:

\begin{equation}
C^{B}_{acc} = accuracy(\delta_{\mathcal{A}}^{d, \tau}, \delta_{\mathcal{L}}^{d, \tau})
\end{equation}

where $accuracy$ denotes the binary accuracy function \cite{Metz1978BasicPO}, and $\delta_{\mathcal{A}}^{d, \tau}$ and $\delta_{\mathcal{L}}^{d, \tau}$ denote $\delta(x, \mathcal{X}, \tau, d)$ and $\delta(z, \mathcal{Z}, \tau, d)$, respectively.

\subsection{Distance Measures}\label{subsec:distance}

The main assessment of this work is based on distance comparisons between original clip fragments and their transformations, both in audio and embedded space. To our best knowledge, not many general ways are developed to calculate the distance between raw audio representations of music signals directly. Therefore, we choose to calculate the distance between audio samples using time-frequency representations as the potential proxy of perceptual distance between the music signals. More specifically, we use Mel Frequency Cepstral Coefficients (MFCCs) with 25 coefficients, dropping the first coefficient when the actual distance is calculated. Eventually, we employ two distance measures on the audio domain:

\begin{itemize}
\setlength\itemsep{0.2em}

\item Dynamic Time Warping (DTW) is a well-known dynamic programming method for calculating similarities between time series. For our experiments, we use the FastDTW implementation~\cite{salvadorfastdtw}.

\item Similarity Matrix Profile (SiMPle)~\cite{DBLP:conf/ismir/SilvaYBK16} measures the similarity between two given music recordings using  a similarity join~\cite{DBLP:conf/ismir/SilvaYBK16}. We take the median of the profile array as the overall distance between two audio signals.
\end{itemize}

For deep embedding space, since any deep representation of input $x$ is encoded as a fixed-length vector $z$ in our models, we adopted two general distance measures for vectors: Euclidean distance and cosine distance.

\subsection{Transformations}\label{subsec:transformations}

\begin{figure}
  \centering
  \includegraphics[width=0.6\textwidth]{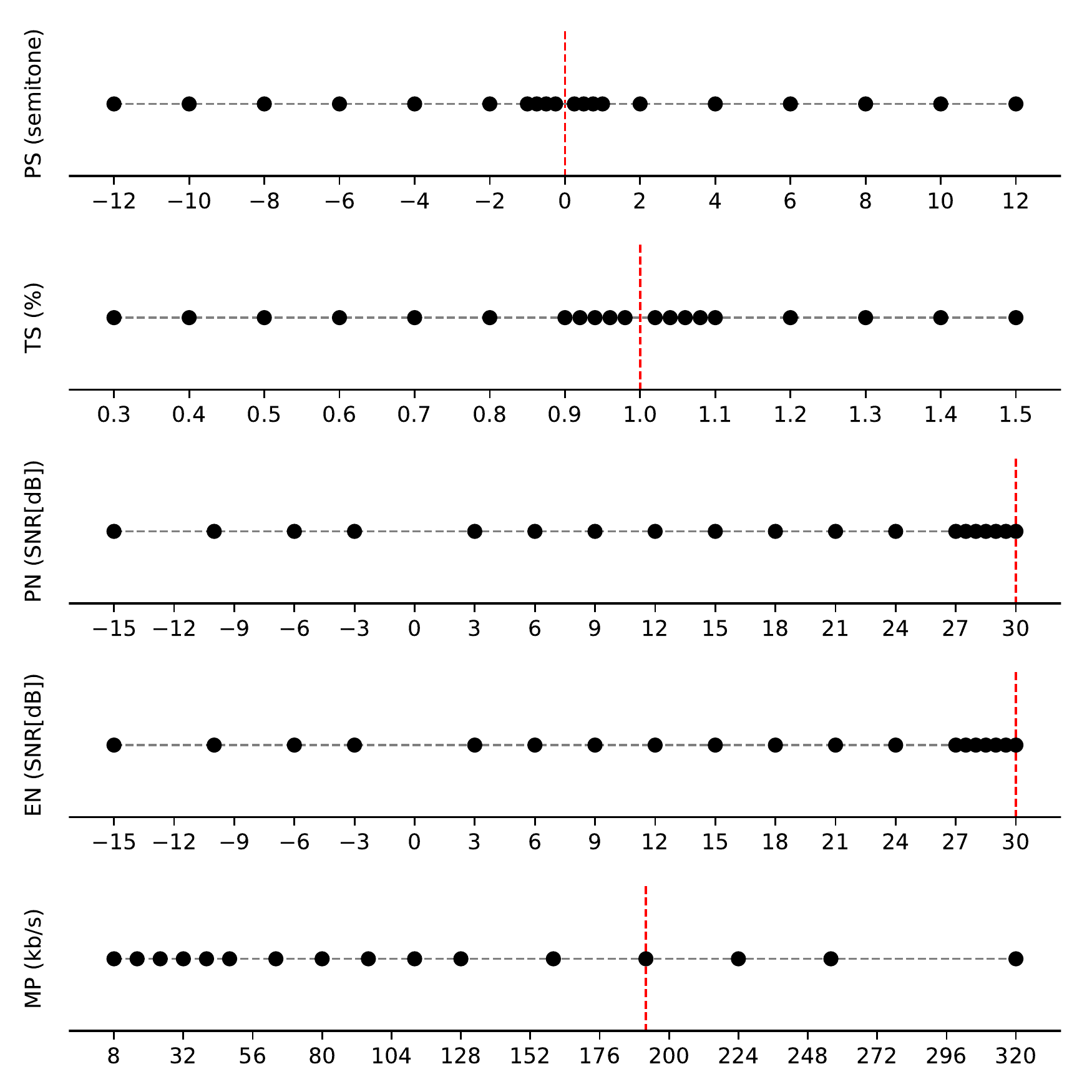}
  \caption{The selected range of magnitudes with respect to the transformations. Each row indicates a transformation category; each dot represents the selected magnitudes. We selected relatively more points in the range in which transformations should have small effect, except for the case of MP3 compression. Here, we tested all the possible transformations (kb/s levels) as supported by the compression software we employed. The red vertical lines indicate the position of the original sample with respect to the transformation magnitudes. For TS and PS, these consider no transformation; for PN, EN and MP, they consider the transformation magnitude that will be closest to the original sample.}
  \label{fig:range}
\end{figure}

In this subsection, we describe the details on the transformations we employed in our experiment. In all cases, we will consider a range from very small, humanly imperceptible transformations, up to transformations within the same category, that should be large enough to become humanly noticeable. While it is not trivial to set an upper bound for the transformation magnitudes, at which a transformed sample may be recognized as a `different' song from the original, we introduce a reasonable range of magnitudes, such that we can investigate the overall robustness of our target encoders as transformations will become more grave. The selected range per each transformation is illustrated in Figure \ref{fig:range}.

\begin{itemize}
\setlength\itemsep{0em}
\item Noise:
As a randomized transformation, we applied both pink noise (PN) and environmental noise (EN) transformations. More specifically, for EN, we used noise recorded in a bar, as collected from \texttt{freesound}.\footnote{\href{https://freesound.org/people/kwahmah_02/sounds/245041/}{https://freesound.org}} The test range of the magnitude, expressed in terms of Signal to Noise Ratio, spans from -15dB to 30dB, with denser sampling for high Signal to Noise Ratios (which are situations in which transformed signals should be very close to the original signal)~\cite{DBLP:journals/corr/KurakinGB16}. This strategy also is adopted for the rest of the transformations.

\item Tempo Shift:
We applied a tempo shift (TS), transforming a signal to a new tempo, ranging from 30\% to 150\% of the original tempo. Therefore, we both slow down and speed up the signal. Close to the original tempo, we employed a step size of 2\%, as a -2\% and 2\% tempo change has been considered as an irrelevant slowdown or speedup in previous work~\cite{DBLP:journals/tmm/Sturm14}. We employed an implementation\footnote{\url{https://breakfastquay.com/rubberband/}} using a phase vocoder and resampling algorithm.

\item Pitch Shift:
We also employed a pitch shift (PS), changing the pitch of a signal, making it lower or higher. Close to the original pitch, we consider transformation steps of $\pm{25}$ cents, which is 50\% smaller than the error bound considered in the MIREX challenge of multiple fundamental frequency estimation \& tracking~\cite{DBLP:conf/ismir/SalamonU12}. Beyond a difference of 1 semitone with respect to the original, whole tone interval steps were considered.

\item Compression:
For compression (MP), we simply compress the original audio sample using an MP3 encoder, taking all kb/s compression rates as provided by the \textit{FFmpeg} software~\cite{tomar2006converting}.

\end{itemize}

For the rest of the paper, for brevity, we use OG as the acronym of the original samples.

\section{Experiment}\label{sec:experiment}

\subsection{Audio Pre-processing}\label{subsec:preproc}

For the input time-frequency representation to the DNNs, we use the dB-scale magnitude STFT matrix. For the calculation, the audio was resampled at 22,050 kHz. The window and overlap size are 1,024 and 256 respectively. It leads to the dimensionality of the frequency axis to be $b=513$, only taking positive frequencies into account. The standardization over the frequency axis is applied by taking the mean and the standard deviation of all magnitude spectra in the training set.

Also, we use the short excerpts of the original input audio track with $t=128$, which yields approximately 2 seconds per excerpt in the setup we used. Each batch of excerpts is randomly cropped from 24 randomly chosen music clips before being served to the training loop.

When applying the transformations, it turned out that some of the libraries we used did not only apply the transformation, but also changed the loudness of the transformed signal. To mitigate this, and only consider the actual transformation of interest, we applied a loudness normalization based on the EBU-R 128 loudness measure~\cite{JHKM:web/ebur128/EBU}. More specifically, we calculated the mean loudness of the original sample, and then ensured that transformed audio samples would have equal mean loudness to their original.

\subsection{Baseline}\label{subsec:baseline}

Beyond deep encoders, we also consider a conventional feature extractor: MFCCs, as also used in~\cite{DBLP:conf/ismir/ChoiFSC17}. The MFCC extractor can also be seen as an encoder, that projects raw audio measurements into a latent embedding space, where the projection was hand-crafted by humans to be perceptually meaningful.

We first calculate the first- and second-order time derivatives of the given MFCCs and then take the mean and standard deviation over the time axis, for the original and its derivatives. Finally, we concatenate all statistics into one vector. Using the 25 coefficients excluding the first coefficient, we obtain $z^{MFCC}\in\mathbb{R}^{144}$ from all the points in $\mathcal{X}^{ts}$. For the AT task, we trained a dedicated $h$ for auto-tagging, with the same objective as Eq.~\ref{eq:at_loss}, while $f$ is substituted as $z^{MFCC}$.

\subsection{Dataset}\label{subsec:dataset}

We use a subset of the Million Song Dataset (MSD)~\cite{DBLP:conf/ismir/Bertin-MahieuxEWL11} both for training and testing of AT and AE task. The number of the training samples $\vert\mathcal{X}^{tr}\vert$ is 71,512. These are randomly drawn from the original subset of 102,161 samples without replacement. For the test set $\mathcal{X}^{ts}$, we used 1,000 excerpts randomly sampled from 1,000 preview clips which are not used at training time. As suggested in~\cite{choi2017convolutional}, we used the top $K=50$ social tags based on their frequency within the dataset.

As for the IR task, we choose to use the training set of the IRMAS dataset~\cite{DBLP:conf/ismir/BoschJFH12}, which contains 6,705 audio clips of 3-second polyphonic mixtures of music audio, from more than 2,000 songs. The pre-dominant instrument of each short excerpt is labeled. As excerpts may have been clipped from a single song multiple times, we split the dataset into training, validation and test sets at the song level, to avoid unwanted bleeding among splits.

Finally, for VS, we employed the MUSDB18 dataset~\cite{musdb18}. This dataset is developed for musical blind source separation tasks, and has been used in public benchmarking challenges~\cite{DBLP:conf/ica/StoterLI18}. The dataset consists of 150 unique full-length songs, both with  mixtures and isolated sources of selected instrument groups: \emph{vocals}, \emph{bass}, \emph{drums} and \emph{other}. Originally, the dataset is split into a training and test set; we split the training set into a training and validation set (with a 7:3 ratio), to secure validation monitoring capability.

Note that since we use different datasets with respect to the tasks, the measurements we investigate will also depend on the datasets and tasks. However, across tasks, we always use the same encoder architecture, such that comparisons between tasks can still validly be made.

\subsection{Performance Measures}\label{subsec:measures}

As introduced in Section \ref{sec:measuring}, we use distance consistency measures as primary evaluation criterion of our work. Next to this, we also measure the performance per employed learning task. For the AE task, the 
Mean Square Error (MSE) is used as a measure of reconstruction error. For the AT task, we apply a measure derived from the popular Area Under ROC Curve (AUC): more specifically, we apply $AUC^C$, averaging the AUC measure over clips. As for the IR task, we choose to use accuracy. Finally, as for the VS task, we choose to use the Signal to Distortion Ratio (SDR), which is one of the evaluation measures used in the original benchmarking campaign. For this, we employ the public software as released by the benchmark organizers. While beyond SDR, this software suite also can calculate 3 more evaluation measures (Image to Spatial distortion Ratio (ISR), Source to Interference Ratio (SIR), Sources to Artifacts Ratios (SAR)), in this study, we choose to only employ SDR: the other metrics consider spatial distortion, while this is irrelevant to our experimental setup, in which we only use mono sources.

\section{Results}\label{sec:result}

In the following subsections, we present the major analysis results for \textit{task-specific performance}, \textit{within-space consistency}, and finally, \textit{between-space consistency}. Shared conclusions and discussions following from our observations will be presented in Section~\ref{sec:disc_concl}.

\subsection{Task-Specific Performance}\label{subsec:task_measure}

\begin{figure}[h]
\centering
  \includegraphics[width=\textwidth]{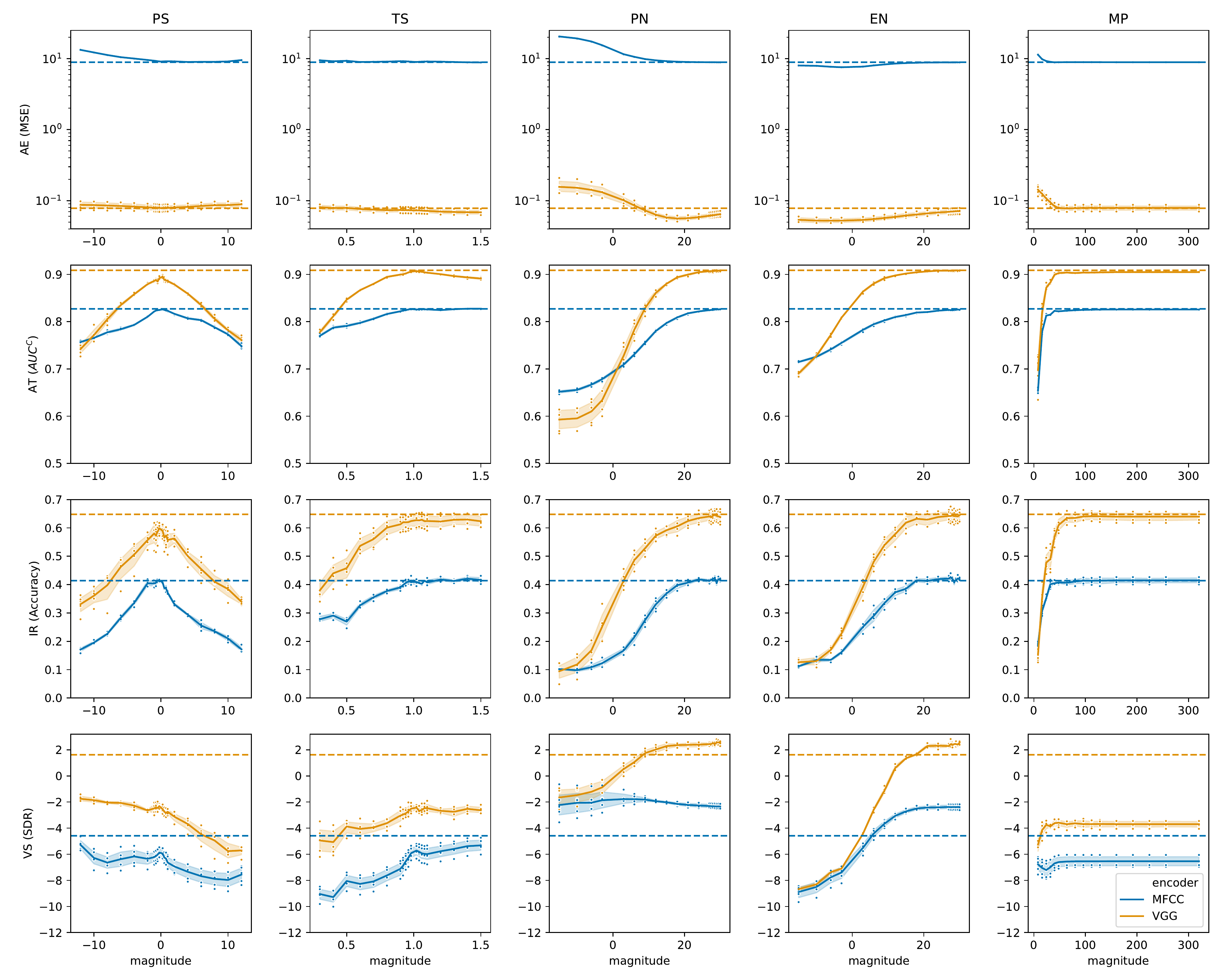}
  \caption{Task specific performance results. Blue and yellow curves indicate the performance of different encoders for each task, over the range of magnitude with respect to the transformations. The performance of original samples is indicated as dotted horizontal lines. For the remaining of the paper including this figure, all the confidence intervals are computed with 1,000 bootstraps at the 95\% level.}
  \label{fig:task_performance}
\end{figure}

To analyze task-specific performance, we ran predictions for the original samples in $\mathcal{X}^{ts}$, as well as their transformations using all $\tau\in\mathcal{T}$ with all the magnitudes we selected. The overall results, grouped by transformation, task and encoder, are illustrated in Figure \ref{fig:task_performance}.
For most parts, we observe similar degradation patterns within the same transformation type. For instance, in the presence of PN and EN transformations, performance decreases in a characteristic non-linear fashion as more noise is added. The exception seems to be the AE task, which shows somewhat unique trends with a more distinct difference between encoders. In particular, when EN is introduced, performance increases with the severity of the transformation. This is likely to be caused by the fact that the environmental noise that we employed is semantically irrelevant for the other tasks, thus causing a degradation in performance. However, because the AE task just reconstructs the given input audio regardless of the semantic context, and the environmental noise that we use is likely not as complex as music or pink noise, the overall reconstruction gets better.

To better understand the effect of transformations, we fitted a Generalized Additive Model (GAM) on the data, using as predictors the main effects of the task, encoder and transformation, along with their two-factor interactions. Because the relationship between performance and transformation magnitude is very characteristic in each case, we included an additional spline term to smooth the effect of the magnitude for every combination of transformation, task and encoder. In addition, and given the clear heterogeneity of distributions across tasks, we standardized performance scores using the within-task mean and standard deviation scores. Furthermore, MSE scores in the AE task are reversed, so that higher scores imply better performance. The analysis model explains most of the variability ($R^2=.98$).

\begin{figure}[h]
\centering
  \includegraphics[width=0.6\textwidth]{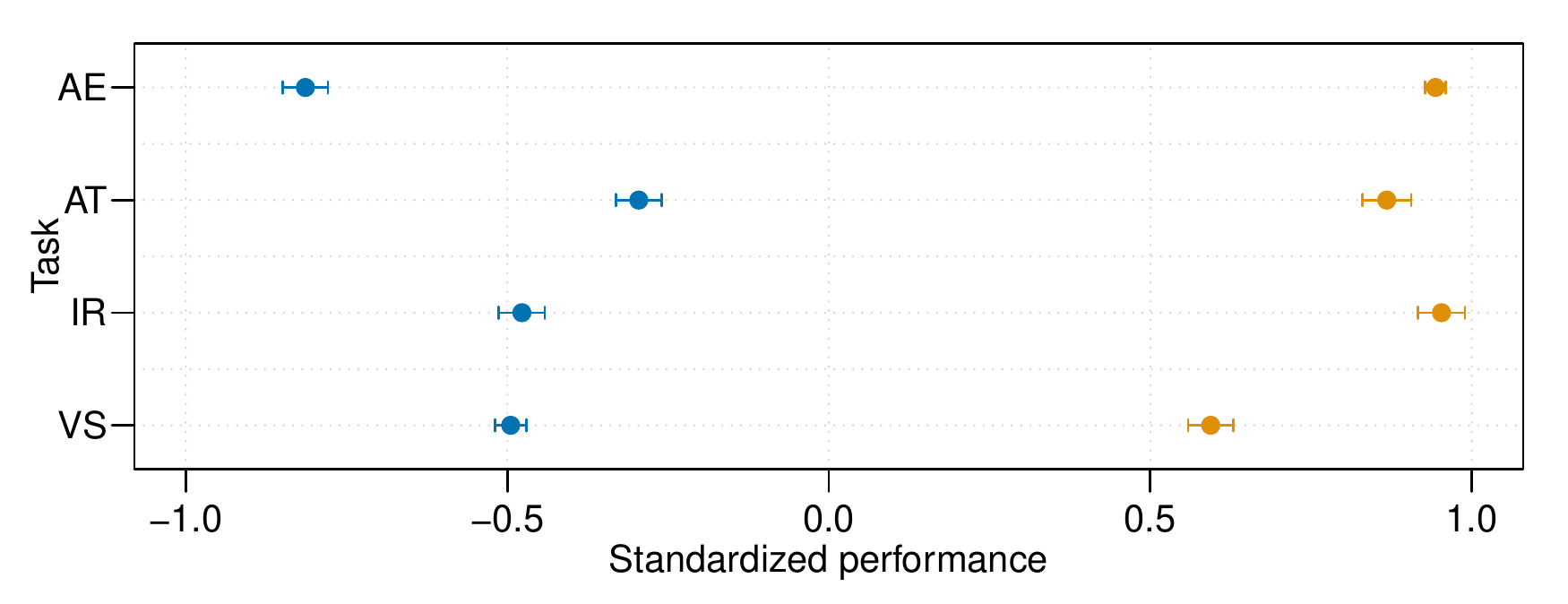}
  \caption{Estimated marginal mean of standardized performance by encoders and tasks, with 95\% confidence intervals. Blue points and brown points indicate the performance of MFCC and VGG-like, respectively.}
  \label{fig:task_performance_emmean}
\end{figure}

An Analysis on Variance (ANOVA) using the marginalized effects clearly reveals that the largest effect is due to the encoders ($F(1, 3522)=12898, p<.0001$), as evidenced by Figure \ref{fig:task_performance}. Indeed, the VGG-like network has an estimated mean performance of $0.84\pm .008$ ($mean\pm s.e.$) standardized units, while MFCCs has an estimated performance of $-0.52\pm .009$ standardized units. The second largest effect is the interaction between transformation and task ($F(12, 3522)=466, p<.0001$), mainly because of the VS task. Comparing the VGG-like and MFCC encoders on the same task ($F(3, 3522)=210, p<.0001$), the largest performance differences appear in the AE task, with VS showing the smallest differences. It suggests that MFCCs loses a substantial amount of information required for reconstruction, while a neural network is capable of maintaining sufficient information to do a reconstruction task. The smallest performance differences in the VS task mostly relate to the performance of the VGG-like encoder, that shows substantial performance degradation in response to the transformations. Figure \ref{fig:task_performance_emmean} shows the estimated mean performance.

\subsection{Within-Space Consistency}\label{subsec:within_space}

\begin{figure}[h]
\centering
  \includegraphics[width=\textwidth]{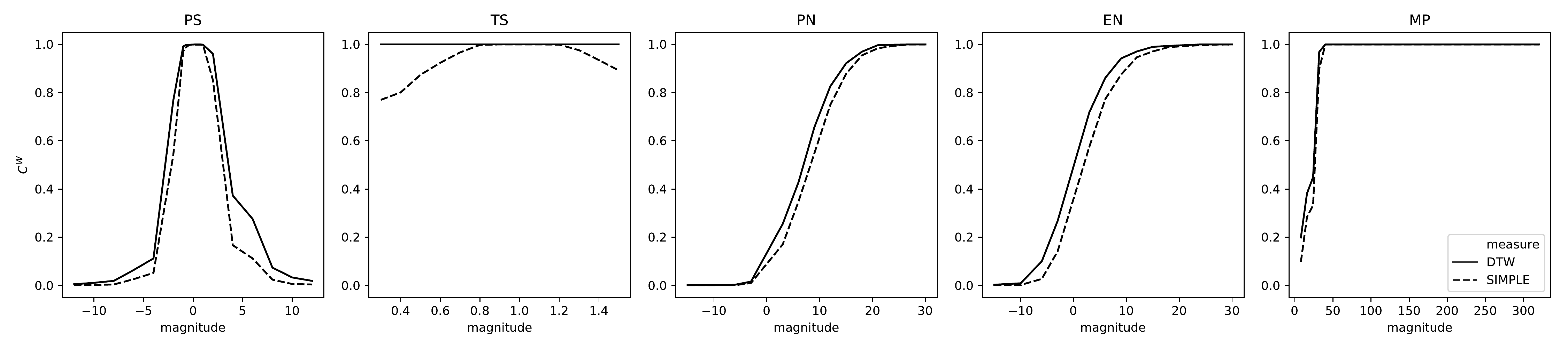}
  \caption{Within-space consistency by transformation on the audio space $\mathcal{A}$. Each curve indicates the within-space consistency $C^{W}$.}
  \label{fig:within_x}
\end{figure}

In terms of within-space consistency, we first examine the original audio space $\mathcal{A}$. As depicted in Figure \ref{fig:within_x}, both the DTW and SiMPle measures show very high consistency for small transformations. As transformations have higher magnitude, as expected, the consistency decreases, but at different rates, depending on the transformation. The clear exception is the TS transformation, where both measures, and in particular DTW, are highly robust to the magnitude of the shift. This result implies that the explicit consideration of both measures on the temporal dynamics can be beneficial.

\begin{figure}[h]
\centering
  \includegraphics[width=\textwidth]{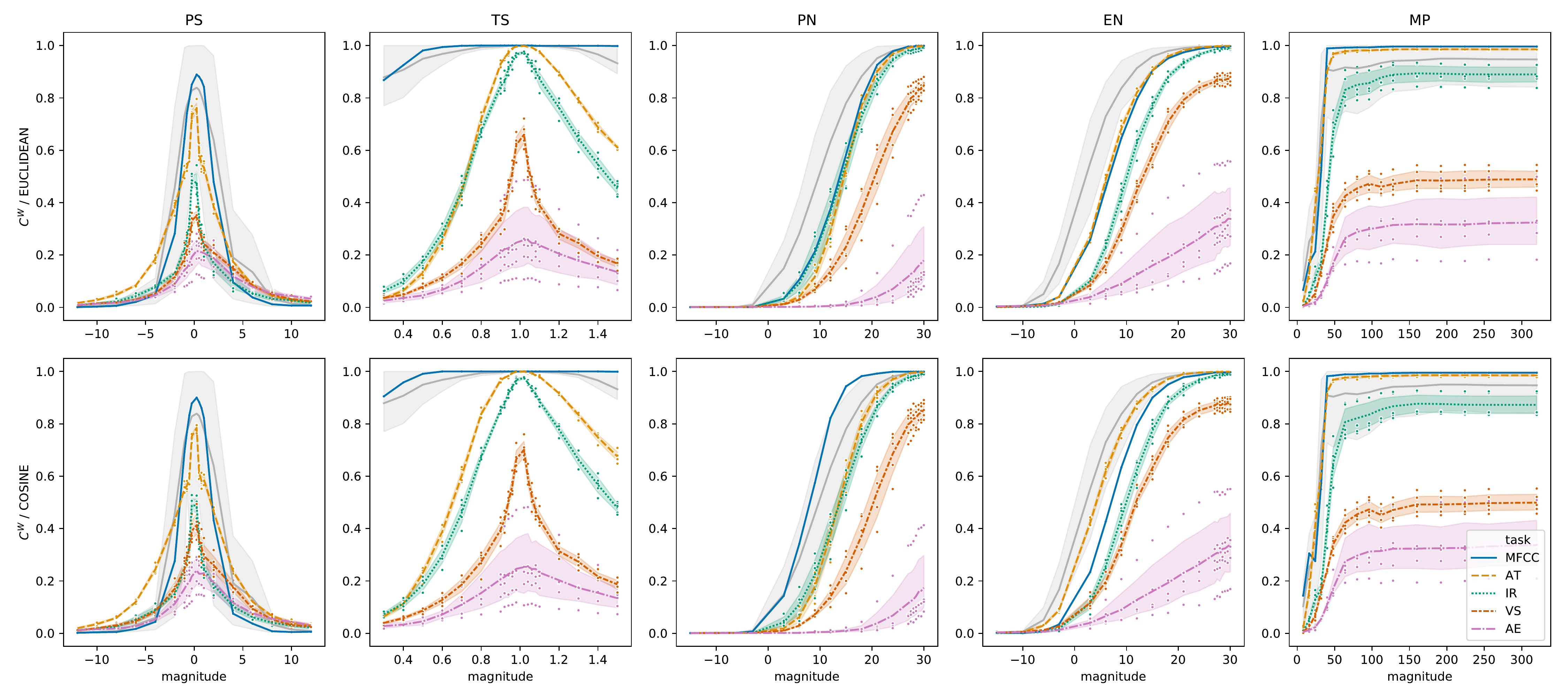}
  \caption{Within-space consistency by transformation on the latent space $\mathcal{L}$. Each curve indicates the within-space consistency $C^{W}$ by task and transformation. The gray curves indicate $C^W$ on $\mathcal{A}$, taken as a weak upper bound for the consistency in the latent space. Confidence intervals are drawn at the 95\% level. Points indicate individual observations from different trials.}
  \label{fig:within_z}
\end{figure}

\begin{figure}[h]
    \centering
    \includegraphics[width=\textwidth]{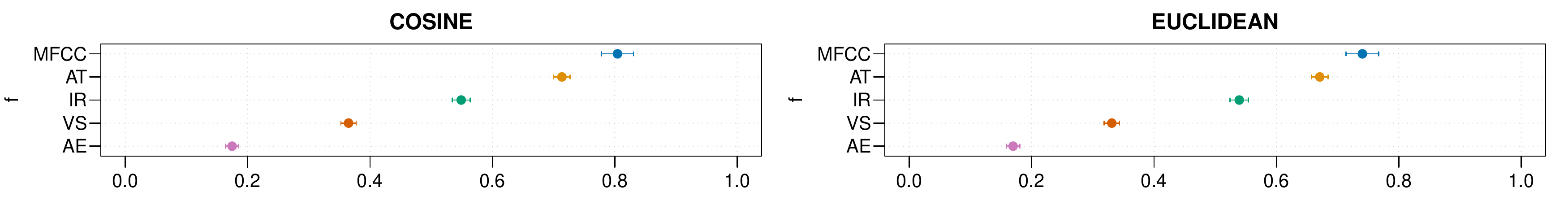}
    \caption{Estimated marginal mean within-space consistency $C^{W}$ in the latent domain. Confidence interval are at 95\% level.}
    \label{fig:within_z_emmean}
\end{figure}

With respect to the within-consistency of the latent space, Figure \ref{fig:within_z} and \ref{fig:within_z_emmean} depicts the results for both the Euclidean and cosine distance measures. In general, the trends are similar to those found in Figure \ref{fig:within_x}. For analysis, we fitted a similar GAM model, including the main effect of the transformation and task, their interaction, and a smoother for the magnitude of each transformation within each task. When modeling consistency with respect to Euclidean distance, this analysis model achieved $R^2=.98$. An ANOVA analysis shows very similar effects due to transformation ($F(4,1793)=1087, p<.0001$) and due to tasks ($F(4,1793)=1066, p<.0001$), with a smaller effect of the interaction. In particular, the model confirms the observation from the plots that the MFCC encoder has significantly higher consistency ($0.741\pm .014$) than the others. For the VGG-like cases, AT shows the highest consistency ($0.671\pm .007$), followed by IR ($0.539\pm .008$), VS ($0.331\pm .007$) and lastly by AE ($0.17\pm .006$). As Figure \ref{fig:within_z_emmean} shows, all these differences are statistically significant.

A similar model to analyze consistency with respect to the cosine distance yielded very similar results ($R^2=0.981$). However, the effect of the task ($F(4,1794)=1263, p<.0001$) was larger than the effect of the transformation ($F(4,1794)=913, p<.0001$), indicating that the cosine distance is slightly more robust to transformations than the Euclidean distance.

\begin{figure*}
\centering
  \includegraphics[width=0.9\textwidth]{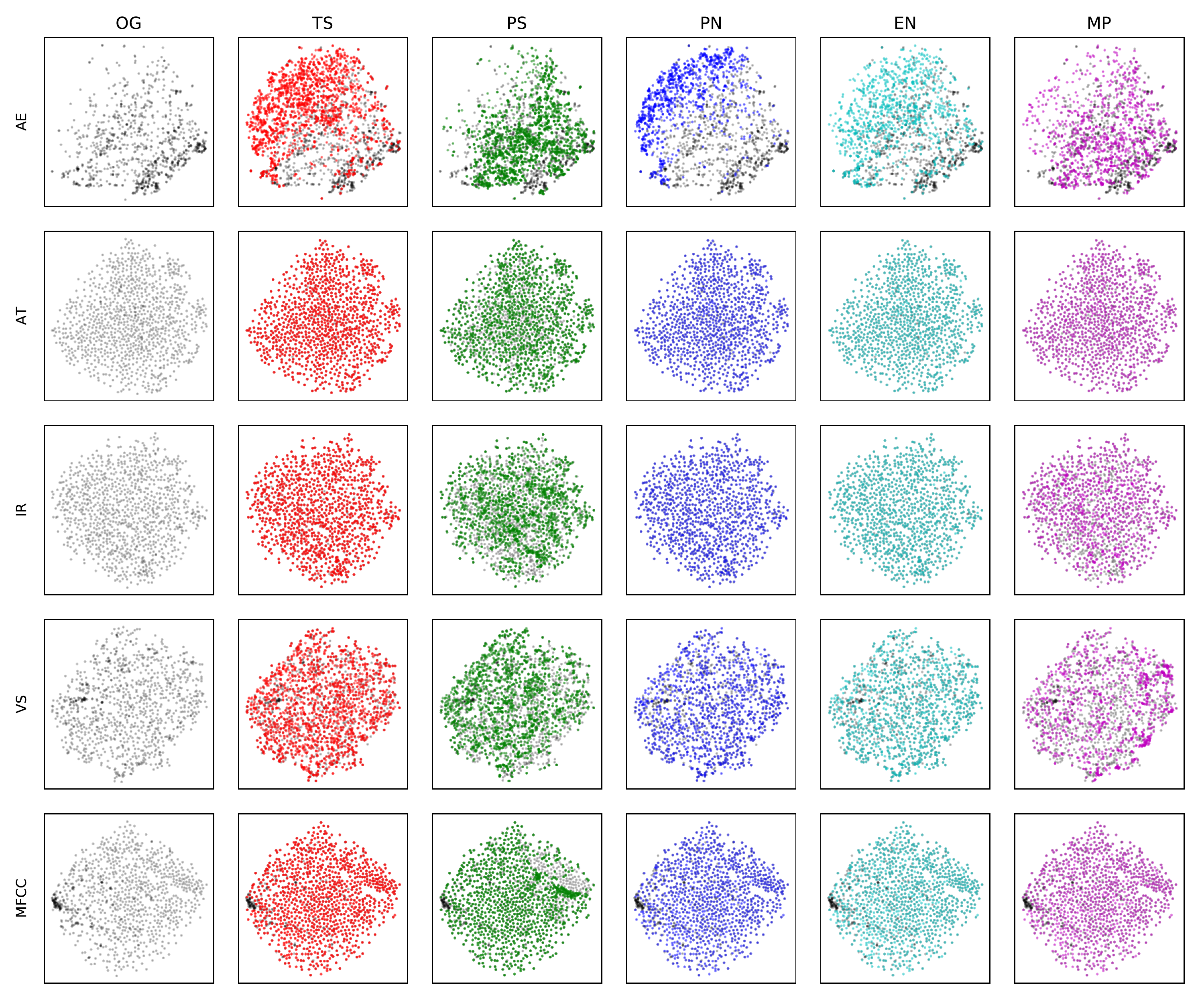}
  \caption{Scatter plot of encoded representations and their transformations for baseline MFCC and $f$ encoders with respect to the tasks we investigated. For all panes, black points indicate original audio samples in the encoded space, and the colored, overlaid points indicate the embeddings of transformations according to the indicated category.}
  \label{fig:tsne}
\end{figure*}

To investigate observed effects more intuitively, we visualize in Figure \ref{fig:tsne} the original dataset samples and their smallest transformations, which should be hardly perceptible to imperceptible to human ears~\cite{DBLP:journals/tmm/Sturm14, DBLP:journals/tmm/KereliukSL15, DBLP:conf/ismir/SalamonU12}\footnote{The smallest transformations are $\pm25$ cents in PS, $\pm2\%$ in TS, 30dB in PN and EN, and 192 kb/s in MP.} in a 2-dimensional space, using t-SNE~\cite{maaten2008visualizing}.
In MFCC space, (Figure~\ref{fig:tsne}), the distributions of colored points, corresponding to each of the transformation categories, are virtually identical to those of the original points. This matches our assumption that very subtle transformations, that humans will not easily recognize, should stay very close to the original points. Therefore, if the hidden latent embedded space had high consistency with respect to the audio space, the distribution of colored points should be virtually identical to the distribution of original points. However, this is certainly not the case for neural networks, especially for tasks such as AE and VS (see~Figure~\ref{fig:tsne}). For instance, in the AE task every transformation visibly causes clusters that do not cover the full space. This suggests that the model may recognize transformations as important \textit{features}, characterizing a subset of the overall problem space.

\subsection{Between-Space Consistency}\label{subsec:between_space}

Next, we discuss between-space consistency according to $C^{B}_{acc}$ and $C^{B}_{\rho}$, as discussed in Section~\ref{subsubsec:between_space}. As in the previous section, we first provide a visualization of the relationship between transformations and consistency, and then employ the same GAM model to analyze individual effects. The analysis will be presented for all pairs of distance measures and between-space consistency measures, which results in 4 models for $C^{B}_{acc}$ and another 4 models for $C^{B}_{\rho}$. As in the within-space consistency analysis, we set the MFCC and other VGG-like networks from different learning tasks as independent `encoder' $f$ to a latent embedded space.

\subsubsection{Accuracy: $C^{B}_{acc}$}\label{subsubsec:acc_X2Z}

\begin{figure}[h]
	\centering
    \includegraphics[width=\textwidth]{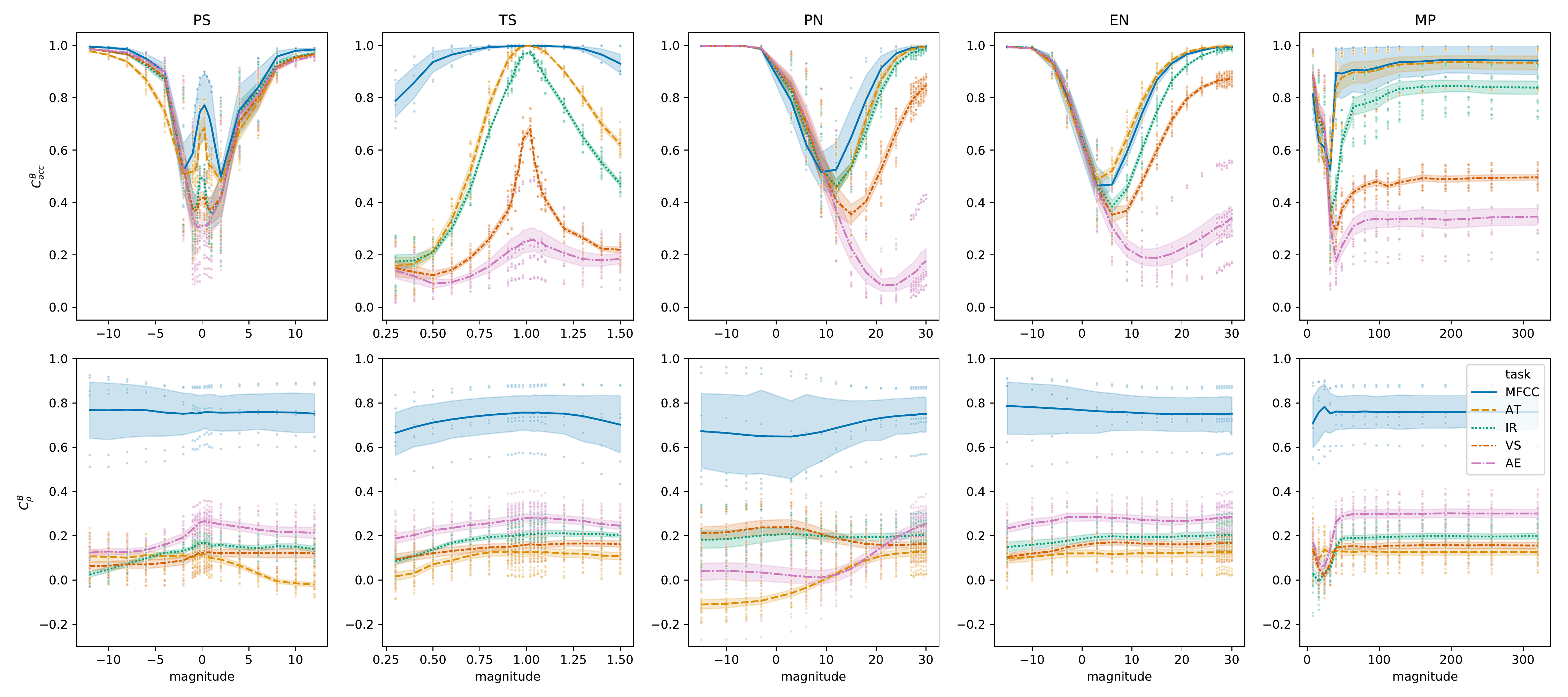}
    \caption{$C^{B}_{acc}$ (top) and $C^{B}_{\rho}$ (bottom) between-space consistency by transformation and magnitude. Each curve indicates the between-space consistency $C^{B}$ with respect to the task. Confidence intervals are drawn at the 95\% level. Points indicate individual observations from different trials.}
	\label{fig:between_space_trans_space}
\end{figure}

The between-space consistency, according to the $C^{B}_{acc}$ criterion, is plotted in the upper plots of Figure~\ref{fig:between_space_trans_space}. Comparing this plot to the within-space consistency plots for $\mathcal{A}$ (Figure~\ref{fig:within_x}) and $\mathcal{L}$ (Figure~\ref{fig:within_z_emmean}), one trend is striking: when within-space consistency in $\mathcal{A}$ and $\mathcal{L}$ becomes substantially low, the between-space consistency $C^{B}_{acc}$ becomes high. This can be interpreted: when grave transformations are applied, the within-space consistencies in both $\mathcal{A}$ and $\mathcal{L}$ space will converge to $0$, and comparing the two spaces, this behavior is consistent.

A first model to analyze the between-space consistency with respect to the SiMPle and cosine measures ($R^2=.96$), reveals that the largest effect is that of the task/encoder $F(4,1772)=440, p<.0001$), followed by the effect of the transformation ($F(4,1772)=285, p<.0001$). The left plot of the first row in Figure \ref{fig:c_b} confirms that the estimated consistency of the MFCC encoder ($0.796\pm .015$) is significantly higher than that of the VGG-like alternatives, which range between $0.731$ and $0.273$. In fact, the relative order is the same as observed in the within-space case: MFCC is followed by AT, IR, VS, and finally AE.

We separately analyzed the data with respect to the other three combinations of measures, and found very similar results. The largest effect is due to the task/encoder, followed by the transformation; the effect of the interaction is considerably smaller. As the first rows of Figure \ref{fig:c_b} shows, the same results are observed in all four cases, with statistically significant differences among tasks.

\subsubsection{Correlation: $C^{B}_{\rho}$}\label{subsubsec:rho_X2Z}

The bottom plots in Figure \ref{fig:between_space_trans_space} show the results for between-space consistency measured with $C^{B}_{\rho}$. It can be clearly seen that MFCC preserves the consistency between spaces much better than VGG-like encoders, and in general, all encoders are quite robust to the magnitude of the perturbations.

\begin{figure}
\centering
\includegraphics[width=\textwidth]{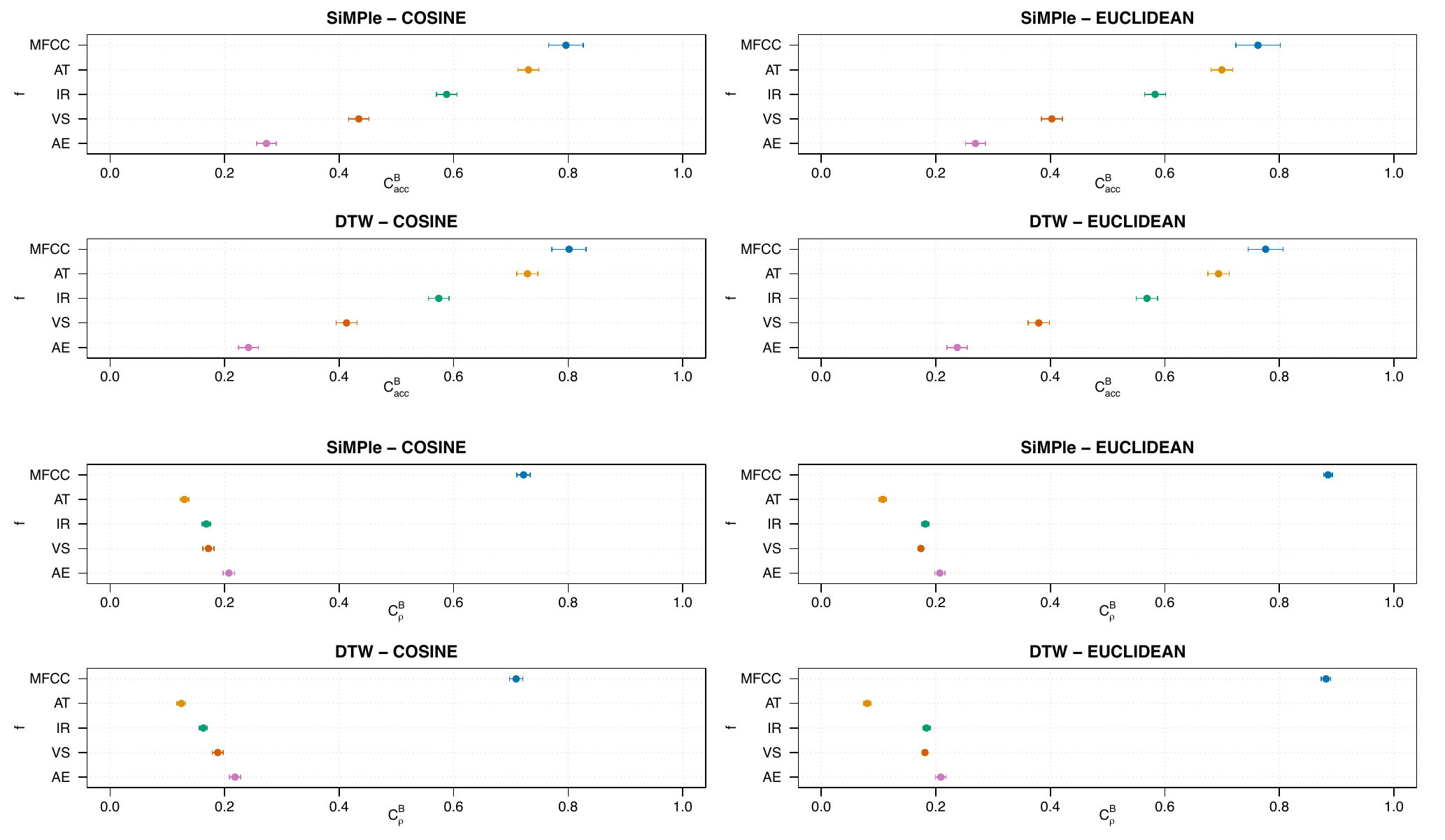}
\caption{Estimated marginal means for between-space consistency by encoder $f$. The first and second rows are for $C^{B}_{acc}$ and the third and fourth rows are for $C^{B}_{\rho}$. Confidence intervals are at the 95\% level.}
\label{fig:c_b}
\end{figure}

Analyzing data again using a GAM model confirms these observations. For instance, when analyzing consistency with respect to the DTW and Euclidean measures ($R^2=0.96$), the largest effect is by far that of the task/encoder ($F(4,1877)=6549,p<.0001$), with the transformation and interaction effect being two orders of magnitude smaller. This is because of the clear superiority of MFCC, with an estimated consistency of $0.881\pm .004$, followed by AE ($0.209\pm .005$), IR ($0.184\pm.003$), VS ($0.181\pm.002$) and finally AT ($0.08\pm .003$) (see right plot of the fourth row in \ref{fig:c_b}).

As before, we separately analyzed the data with respect to the other three combinations of measures, and found very similar results. As first two rows of Figure \ref{fig:c_b} shows, the same qualitative observations can be made in all four cases, with statistically significant differences among tasks. Noticeably, the superiority of MFCC is even clearer when employing the Euclidean distance. Finally, another visible difference is that the relative order of VGG-like networks is reversed with respect to $C^B_{acc}$, with AE being the most consistent, followed by VS, IR, and finally AT.

\subsection{Sensitivity to Imperceptible Transformations}\label{subsec:imperceptible}

\subsubsection{Task-Specific Performance}\label{subsubsec:imperceptible:performance}
\begin{figure}
    \centering
    \includegraphics[width=.9\textwidth]{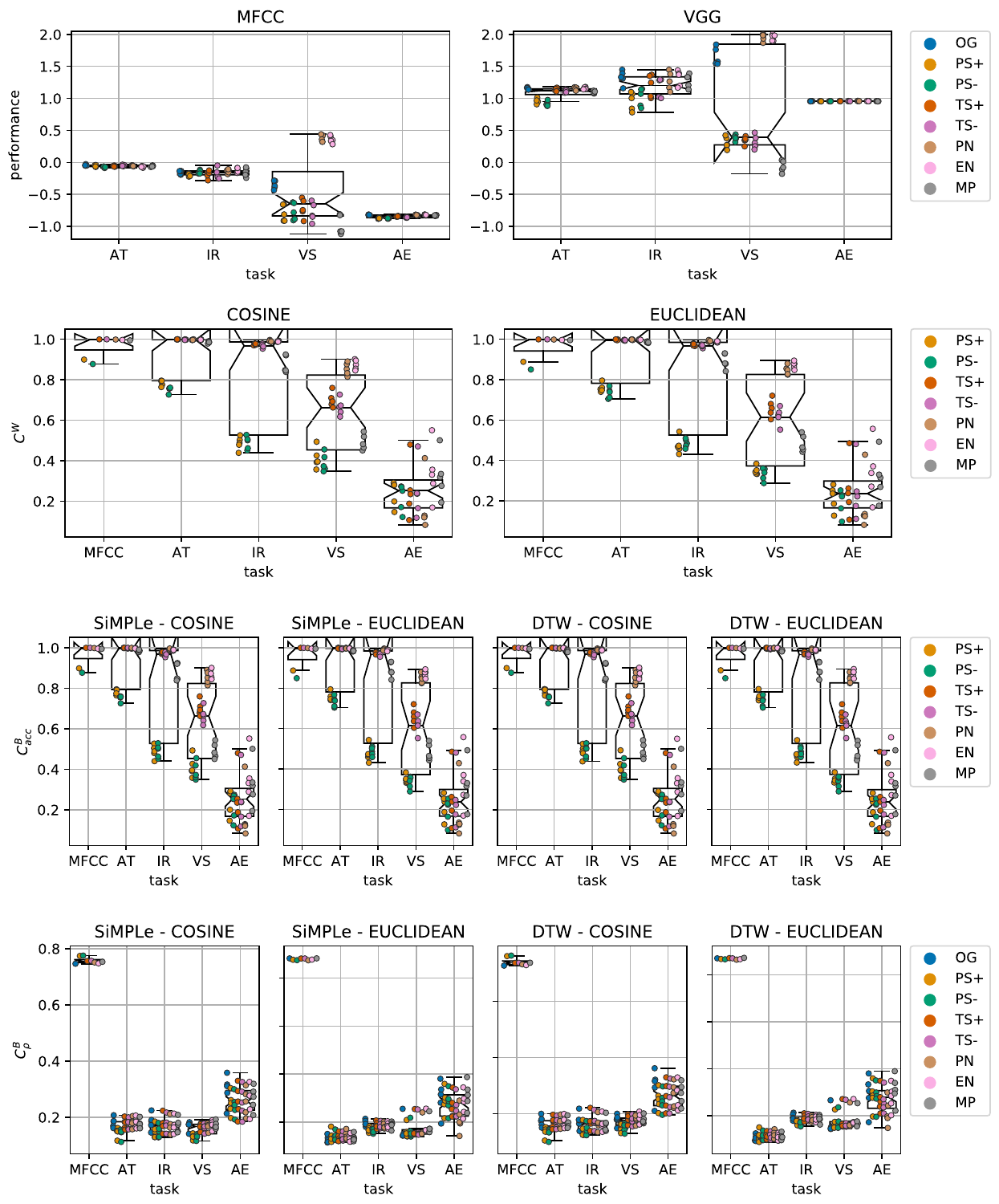}
    \caption{Performance, within-space consistency, and between-space consistency distribution on the minimum transformations. The points are individual observations with respect to the transformation types. For PS and TS, we distinguish in the direction of the transformation (+: pitch/tempo up, -: pitch/tempo down). The first row indicates the task-specific performance, and the second row depicts the within-space consistency $C^{W}$, and finally, the third and fourth rows show the between-space consistency $C^{B}_{acc}$ and $C^{B}_{\rho}$, respectively. The performance is standardized per task, and the sign of AE performance is flipped, similarly to our analysis models.}
    \label{fig:perceptible}
\end{figure}

In this subsection, we focus more on the special cases of transformations with a magnitude small enough to hardly be perceivable by humans~\cite{DBLP:journals/tmm/Sturm14, DBLP:journals/tmm/KereliukSL15, DBLP:conf/ismir/SalamonU12} As the first row of Figure \ref{fig:perceptible} shows, performance is degraded even with such small transformations, confirming the findings from \cite{DBLP:journals/tmm/Sturm14}. In particular, the VS task shows more variability among transformations compared to other tasks. Between transformations, the PS cases show relatively higher degradation.

\subsubsection{Within-Space Consistency}\label{subsubsec:imperceptible:within_z}

The second row of Figure \ref{fig:perceptible} illustrates the within-space consistency on the $\mathcal{L}$ space when considering these smallest transformations. As before, there is no substantial difference between the distance metrics. In general, the MFCC, AT, and IR encoder/tasks are relatively robust on these small transformations, with their median consistencies close to 1. However, encoders trained on the VS and AE tasks show undesirably high sensitivity to these small transformations. In this case, the effect of the PS transformations is even more clear, causing considerable variance for most of the tasks. The exception is AE, which is more uniformly spread in the first place.  

\subsubsection{Between-Space Consistency}\label{subsubsec:imperceptible:between}

Finally, the between-space consistencies on the minimum transformations are depicted in the last two rows of Figure \ref{fig:perceptible}. First, we see no significant differences between pairs of distance measures. When focusing on $C^{B}_{acc}$, the plots highly resemble those from \ref{subsubsec:imperceptible:within_z}, which can be expected, because the within-space consistency on $\mathcal{A}$ is approximately 1 for all these transformations, as illustrated in Figure \ref{fig:within_x}. On the other hand, when focusing on $C^B_\rho$, The last row of Figure \ref{fig:perceptible} shows that even such small transformations already result in large inconsistencies between spaces when employing neural network representations.

\section{Discussion and Conclusion}\label{sec:disc_concl}

\subsection{Effect of the Encoder}\label{sec:disc_concl:encoder}

For most of our experiments, the largest differences are found between encoders. As is well-known, the VGG-like deep neural network shows \emph{significantly better task-specific performance} in comparison to the MFCC encoder. However, when considering distance consistency, MFCC is shown to be \emph{the most consistent encoder} for all cases, with neural network approaches performing substantially worse in this respect. This suggests that, in case a task requires robustness to potential musical/acoustical deviations in the audio input space, it may be more preferable to employ MFCCs than neural network encoders.

\subsection{Effect of the Learning Task}\label{subsec:disc_concl:task}

Considering the neural networks, our results show that the choice of learning task is the most important factor affecting consistency. For instance, a VGG-like network trained on the AE task seems to preserve the relative distances among samples (high $C^B_\rho$), but individual transformed samples will fall closer to originals that were not the actual original the transformation was applied to (low $C^B_{acc}$). On the other hand, a task like AT yields high consistency in the neighborhood of corresponding original samples (high $C^B_{acc}$), but does not preserve the general structure of the audio space (low $C^B_\rho$). This means that a network trained on a low-level task like AE is more consistent than a network trained on a high-level task like AT, because the resulting latent space is less morphed and it more closely resembles the original audio space. In fact, in our results we see that the semantic high-levelness of the task (AT $>$ IR $>$ VS $>$ AE) is positively correlated with $C^B_acc$, while negatively correlated with $C^B_\rho$.

\begin{figure}[!t]
\centering
  \includegraphics[width=0.4\textwidth]{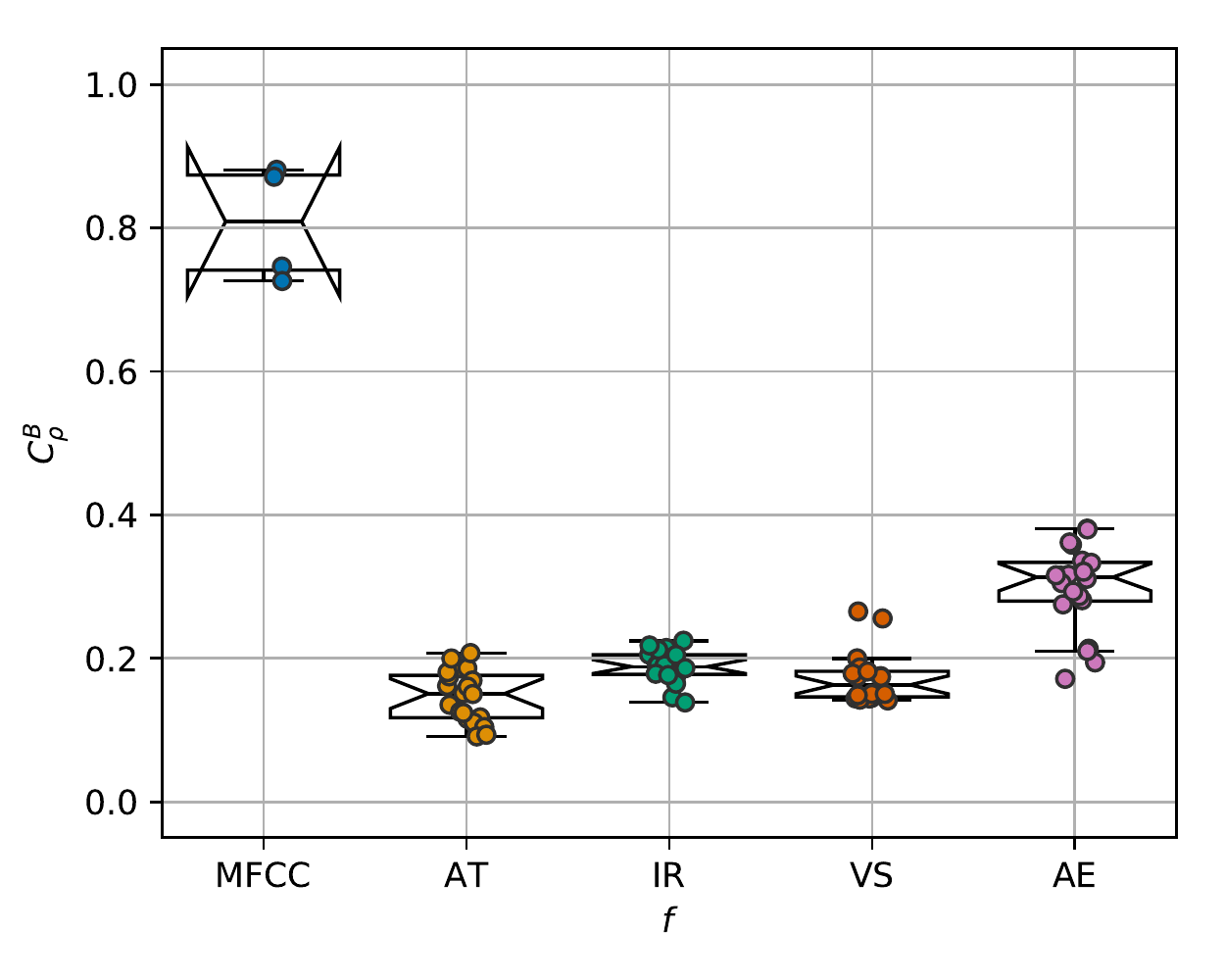}
  \caption{$C^{B}_{\rho}$ on the original samples, including all the possible distance pairs between audio and latent domain.}
  \label{fig:og2og_corr}
\end{figure}

To further confirm this observation, we also computed the between-space consistency $C^{B}_{\rho}$ only on the set of original samples. The results, in Figure \ref{fig:og2og_corr}, are very similar to those in the last two rows of Figure \ref{fig:c_b} and \ref{fig:perceptible}. This suggests that in general, the global distance structure of an embedded latent space with respect to the original samples generalizes over the vicinity of those originals, at least for the transformations that we employed.

\begin{figure}
    \centering
    \includegraphics[width=\textwidth]{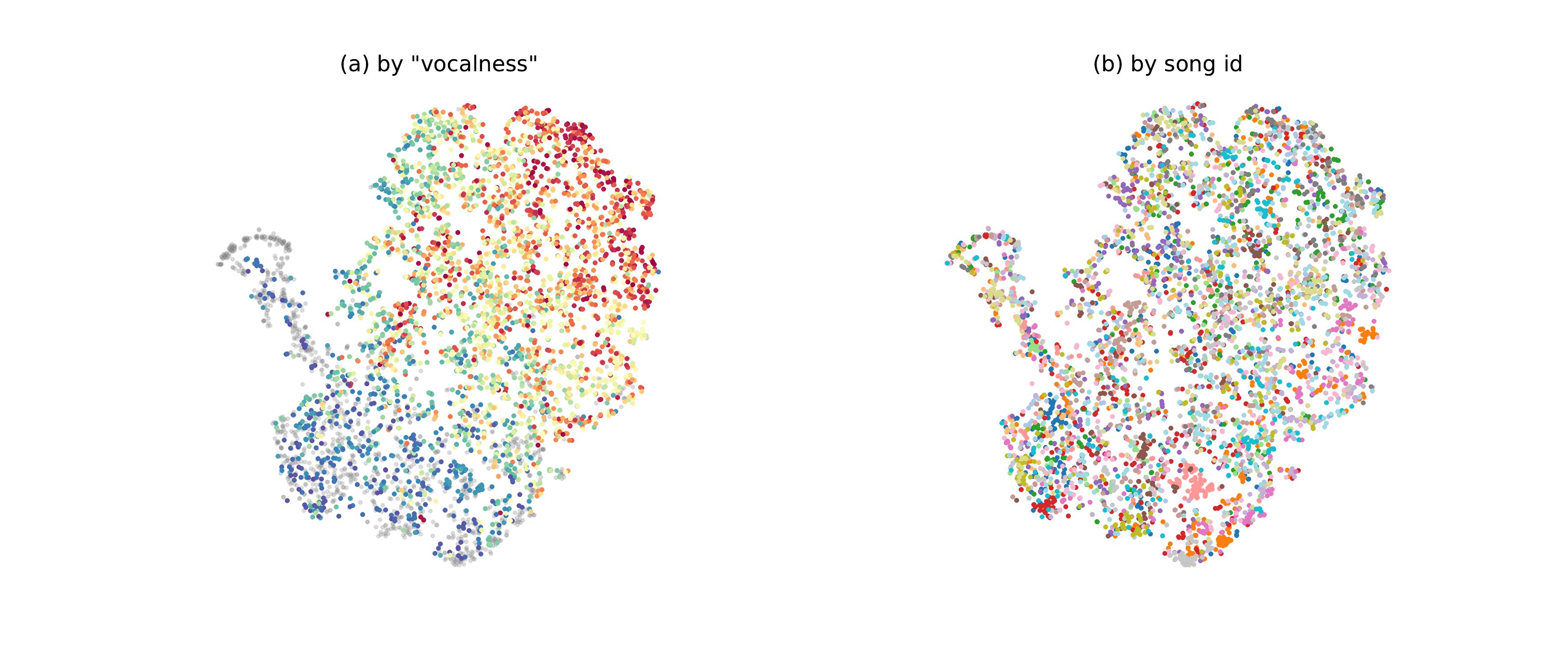}
    \caption{2-dimensional scatter plot using t-SNE. Each point represents 2-second audio mixture signal chunks that are encoded by a VS-specialized encoder. In the left plot, the color map of points is based on the loudness of the isolated vocal signal for a given mixture signal. The red color indicates higher loudness, and the blue color indicates smaller loudness. On the right plot, the same chunks are colored by the song each chunk belongs to. The samples are randomly sampled from the MUSDB18 dataset.}
    \label{fig:vocal_accomp}
\end{figure}

Considering that AE is an unsupervised learning task, and its objective is merely to embed an original data point into a low-dimensional latent space by minimizing the reconstruction error, the odds are lower that data points will cluster according to more semantic criteria, as implicitly encoded in supervised learning tasks. For instance, in contrast, the VS task should morph the latent space such, that input clips with similar degrees of `vocalness' should fall close together, as indeed is shown in Figure \ref{fig:vocal_accomp}. As the task becomes more complex and high-level, such as with AT, this clustering effect will become more multi-faceted and complex, potentially morphing the latent space with respect to the semantic space that is used as the source of supervision. 

\subsection{Effect of the Transformation}\label{subsec:disc_concl:trans}

Across almost all experimental results, significant differences between transformation categories are observed. On the one hand, this supports the findings from~\cite{DBLP:journals/tmm/Sturm14,DBLP:journals/tmm/KereliukSL15}, which show the vulnerability of MIR systems to small audio transformations. On the other hand, this also implies that different types of transformations have different effects on the latent space, as depicted in Figure~\ref{fig:within_z}.

\subsection{Are Nearby Neighbors Relatives?}\label{subsec:disc_concl:z_bad}

As depicted in Figure \ref{fig:within_z}, substantial inconsistencies emerge in $\mathcal{L}$ when compared to $\mathcal{A}$. Clearly, these inconsistencies are not desirable, especially when the transformations we applied are not supposed to have noticeable effects. However, as our consistency investigations showed, the MFCC baseline encoder behaves surprisingly well in terms of consistency, evidencing that hand-crafted features should not always be considered as inferior to deep representations.

While in a conventional audio feature extraction pipeline, important salient data patterns may not be captured due to accidental human omission, our experimental results indicate that DNN representations may be unexpectedly unreliable. In the deep music embedding space,`known relatives' in the audio space may suddenly become faraway pairs. That a representation has certain unexpected inconsistencies should be carefully studied and taken into account, specially given the increasing interest in applying transfer learning using DNN representations, not only in the MIR field. For example, if a system requires to use degraded audio inputs for a pre-trained DNN (which e.g.\ may be done in music identification tasks), while humans may barely recognize the differences between the inputs and their original form, it does not guarantee that this transformed input may be embedded at a similar position to its original version in a latent space.

\subsection{Towards Reliable Deep Music Embeddings}\label{subsec:disc_concl:towards}

In this work, we proposed to use several distance consistency-based criteria, in order to assess whether representations in various spaces can be deemed as consistent. We see this as a complementary means of diagnosis beyond task-related performance criteria, when aiming to learn more general and robust deep representations. More specifically, we investigated whether deep latent spaces are consistent in terms of distance structure, when smaller and larger transformations on raw audio are introduced (\emph{RQ 1}). Next to this, we investigated how various types of learning tasks used to train deep encoders impact the consistencies (\emph{RQ 2}).

Consequentially, we conducted an experiment employing 4 MIR tasks, and considering deep encoders versus a conventional hand-crafted MFCC encoder, to measure the consistency for different scenarios. Our findings can be summarized as follows:

\begin{enumerate}[label=RQ \arabic*.\hspace{0.2cm}, itemindent=4em]
    \item Compared to the MFCC baseline, all DNN encoders indicate lower consistency, both in terms of within-space consistency and between-space consistency, especially when transformations grow from imperceptibly small to larger, more perceptible ones.
    \item Considering learning tasks, the high-levelness of a task is correlated with the consistency of resulting encoder. For instance, an AT-specialized encoder, which needs to deal with semantically high-level task, yields the highest within-space consistency, but the lowest between-space consistency. On the other hand, an AE-specialized encoder, which deals with a semantically low-level task, shows opposite trends.
\end{enumerate}

To realize a fully robust testing framework, there still are a number of aspects to be investigated. First of all, more in-depth study is required considering different magnitudes in the transformations, and their possible comparability. While we applied different magnitudes for each transformations, we decided not to comparatively consider the magnitude ranges in the analysis at this moment. This was done, as we do not have any exact means to compare the perceptual effect of different magnitudes, which will be crucial to regularize between transformations.

Furthermore, similar analysis techniques can be applied to more diverse settings of DNNs, including different architectures, different levels of regularizations, and so on. Also, as suggested in ~\cite{DBLP:journals/tmm/KereliukSL15,DBLP:journals/corr/GoodfellowSS14}, the same measurement and analysis techniques can be used for \textit{adversarial examples} generated from the DNN itself, as another important means of studying a DNN's reliability.

Moreover, and based on the observations from our study, it may be possible to develop countermeasures for maintaining high consistency of a model, while yielding high task-specific performance.
For instance, unsupervised de-noising such as~\cite{DBLP:journals/cee/NazeerBWMANOK18, DBLP:journals/paa/NazeerBJM18} might be one of the potential solutions. In particular, it can be used when the noise is drawn from the known, relatively simple distribution, such as white noise. However, we also observed some encoders are substantially affected by a very small amount of the noise, which implies even artifacts produced from the de-noising algorithm can cause another unexpected inconsistency. Also, it might not guarantee more musical and structured cases such as tempo or pitch shifts.

For those cases, it can be effective if, during learning, a network is directly supervised to treat transformations in similar ways as their original versions in the latent space. This can be implemented as an auxiliary objective to the main objective of the learning procedure, or introducing directly the transformed examples as the data augmentation.

We believe that our work can be a step forward towards a practical framework for more interpretable deep learning models, in the sense that we suggest a less task-dependent measure for evaluating a deep representation, that still is based on known semantic relationships in the original item space.\footnote{The Python code that is used for this experiment can be found in \url{https://github.com/eldrin/are-nearby-neighbors-relatives}}

\section*{Acknowledgments}
We would like to thank Cunquan Qu and Taekyung Kim, for many useful inputs and valuable discussions. This work was carried out on the Dutch national e-infrastructure with the support of SURF Cooperative.

\bibliographystyle{unsrt}  
\bibliography{main}  






\end{document}